\documentclass[11pt]{article}

\usepackage[final]{acl}

\usepackage{times}
\usepackage{latexsym}

\usepackage[T1]{fontenc}

\usepackage[utf8]{inputenc}

\usepackage{multirow}
\usepackage{url}
\usepackage{microtype}

\usepackage{inconsolata}

\usepackage{graphicx}
\usepackage{algorithm}
\usepackage{algorithmic}
\usepackage{amsmath}
\usepackage{amssymb}
\usepackage{booktabs}
\usepackage{times}
\usepackage{helvet}
\usepackage{courier}
\usepackage{xcolor}
\usepackage{enumitem}
\usepackage[table]{xcolor}
\usepackage{subcaption}
\usepackage{listings}
\usepackage{tcolorbox}
\tcbuselibrary{listings}
\tcbuselibrary{skins, breakable}

\title{Disentangling Reasoning Logic to Resolve Explicit Knowledge Conflicts}

\author{
  Xianda Zheng\textsuperscript{1} \quad
  Zijian Huang\textsuperscript{1} \quad
  Meng-Fen Chiang\textsuperscript{2} \quad
  Jiamou Liu\textsuperscript{1} \\
  {\bf Yuan Fang}\textsuperscript{\textbf{3}} \quad
  {\bf Michael Witbrock}\textsuperscript{\textbf{1}} \quad
  {\bf Kaiqi Zhao}\textsuperscript{\textbf{4*}} \\
  \textsuperscript{1}School of Computer Science, University of Auckland\\
  \textsuperscript{2}Department of Electronics and Electrical Engineering, National Yang Ming Chiao Tung University \\
  \textsuperscript{3}School of Computing and Information Systems, Singapore Management University \\
  \textsuperscript{4}Shenzhen Key Laboratory of Internet Information Collaboration, Harbin Institute of Technology (Shenzhen) \\
  \texttt{xzhe162@aucklanduni.ac.nz} \quad
  \texttt{zhaokaiqi@hit.edu.cn} \\
}

\newcommand\blfootnote[1]{%
  \begingroup
  \renewcommand\thefootnote{}\footnote{#1}%
  \addtocounter{footnote}{-1}%
  \endgroup
}

\begin{document}
\maketitle

\blfootnote{$^*$Corresponding author.}

\begin{abstract}
Explicit knowledge conflicts, occurring when retrieved contexts contain contradictory information, pose a fundamental challenge for Large Language Models (LLMs) as they integrate increasingly diverse data sources. The core difficulty lies in the complexity of entangled narratives and heterogeneous conflict patterns, which frequently exceeds the reasoning capacity of standard backbone architectures. We propose \textbf{\textsc{Kcr}} (Knowledge Conflict Reasoning), a framework that adjudicates contradictions by systematically structuring their underlying logic. \textsc{Kcr} disentangles conflicting contexts into discrete sets of reasoning traces, utilizing a hybrid representation of text and graphs to facilitate systematic comprehension. It then employs a Reinforcement Learning with Verifiable Rewards (RLVR) paradigm to instill a reasoning policy that maximizes logical consistency while suppressing spurious paths derived from contradictory evidence. Extensive evaluations demonstrate that \textsc{Kcr} yields substantial performance gains. Notably, a 7B model enhanced by \textsc{Kcr} achieves adjudication capabilities that significantly outperform leading proprietary models, including GPT-4o and GPT-5.1, on complex tasks.
Code is available at \url{https://github.com/zhengxianda/KCR}.
\end{abstract}

\section{Introduction}
\label{sec:introduction}

The integration of knowledge from diverse sources into Large Language Models (LLMs)~\cite{GPT4, llama, qwen} often leads to conflicts, resulting in contradictory conclusions \cite{conflict_survey, benchmark}. For instance, the classification of Pluto as a planet depends on its evolving definition, resulting in conflicting determinations over time. This challenge is exacerbated by the widespread adoption of Retrieval Augmented Generation (RAG), where models often retrieve documents containing opposing claims. Unlike hallucinations generated from internal LLMs' weights, these conflicts originate directly from the context itself. This forces LLMs into the untenable position of adjudicating between contradictory sources without a clear ground truth, leading to unreliable or inconsistent outputs.

\begin{figure}[t] 
    \centering
    \includegraphics[width=\linewidth]{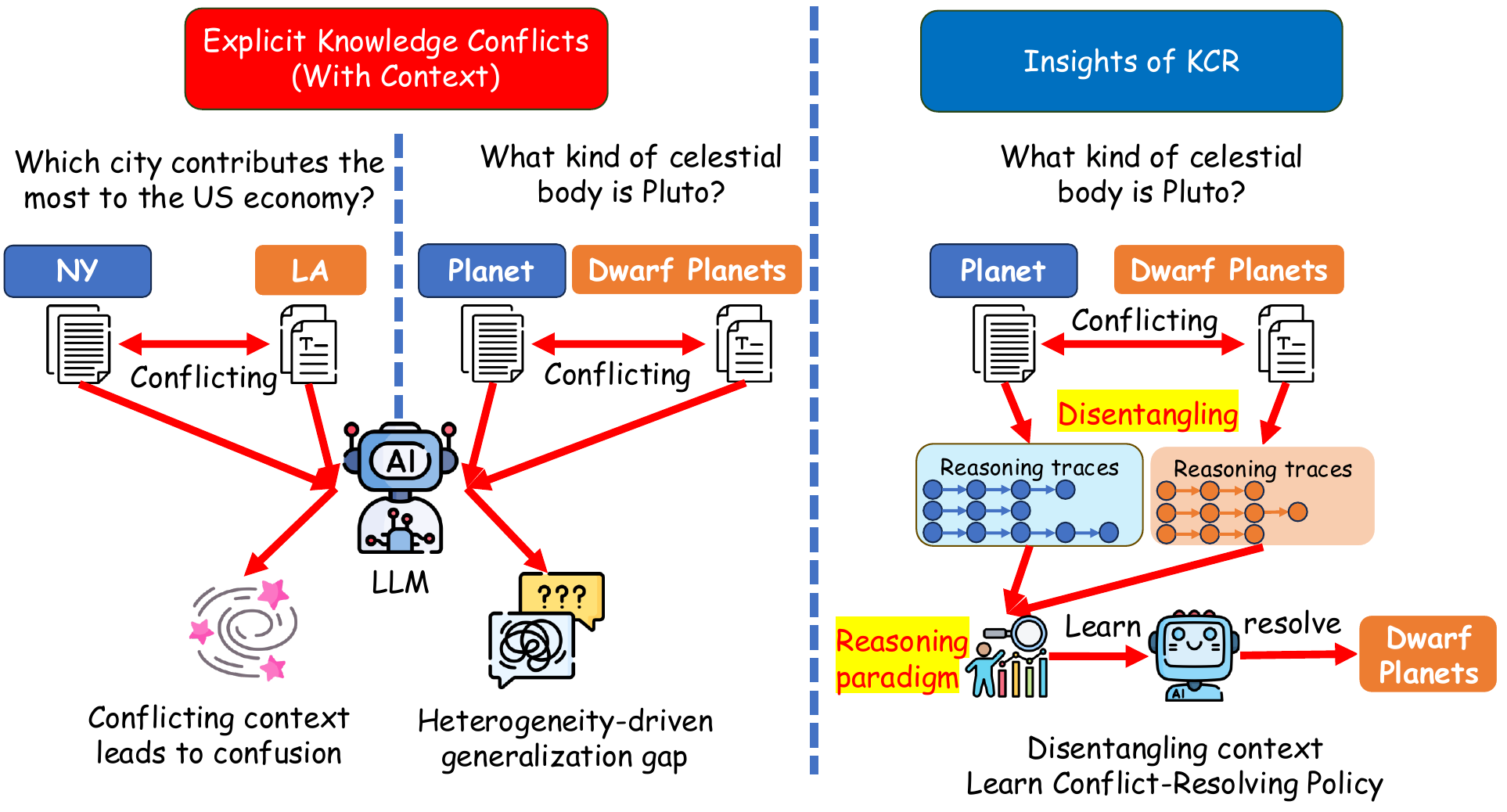} 
    \caption{
     Challenges in resolving explicit knowledge conflicts for LLMs. (a) \textbf{Logic Entanglement}: The interweaving of contradictory evidence within dense contexts impedes the separation of valid reasoning traces, leading to confusion. (b) \textbf{Case Heterogeneity}: Static inductive principles (e.g., fixed rules) fail to generalize across the diverse nature of conflict scenarios, rendering them ineffective for robust adjudication.
    } 
    \label{motivation}
\end{figure}

Prevalent approaches for resolving explicit conflicts typically resort to \textit{knowledge fusion} via semantic decoding \cite{FiD, merging_1}. These methods often employ auxiliary embedding decoders to aggregate arbitrary context \cite{CAD, CUAD}, thereby indirectly resolving explicit knowledge conflicts within the context. 
However, such decoders suffer from significant performance degradation as context length increases. More recent strategies leverage the inherent reasoning capabilities of LLMs to fuse text directly \cite{benchmark}. While promising, current methods often rely on rigid, pre-defined principles or complex multi-round iterative prompting \cite{Astute_RAG}, which are computationally expensive and struggle to generalize across diverse conflict types.

We argue that these limitations persist because existing paradigms fail to address the logic entanglement and case heterogeneity posed by conflicting contexts. 
As illustrated in Figure \ref{motivation}, we identify two primary challenges.
First, \textbf{Logic Entanglement} arises from the dense interweaving of contradictory evidence, which obscures underlying reasoning paths and makes it exceptionally difficult to disentangle valid logic from spurious noise.
Second, \textbf{Case Heterogeneity} renders static inductive approaches (such as fixed principles) ineffective, as they lack the flexibility to generalize across the diverse spectrum of conflict scenarios.
Together, these factors impede standard reasoning processes, necessitating a framework capable of structural disentanglement and robust adjudication.

To address these challenges, we propose \textbf{K}nowledge \textbf{C}onflict \textbf{R}easoning (\textbf{KCR}), a framework designed to internalize a robust conflict-resolution paradigm. \textsc{Kcr} operates in two phases: (i) \textit{Disentangling Conflict Logic} and (ii) \textit{Generalizable Conflict-Resolving Policy Learning}. 
In the first phase, \textsc{Kcr} decomposes entangled contexts into structured logic chains using two complementary formats: textual chains (for unstructured nuance) and local knowledge graphs (for structural clarity). In the second phase, we employ Reinforcement Learning with Verifiable Rewards (RLVR) to align the backbone LLM with a rigorous adjudication process. Specifically, we introduce rewards for \textit{logical coherence} and \textit{consistency}, incentivizing the model to construct valid reasoning chains while actively discriminating against spurious reasoning paths derived from contradictory contexts.

Our contributions are summarized as follows:
\begin{itemize}[leftmargin=*]
\item To the best of our knowledge, \textsc{Kcr} is the first framework designed to resolve Explicit Knowledge Conflicts by structurally disentangling reasoning traces and enhancing the intrinsic adjudication abilities of backbone LLMs.
\item We propose a novel training paradigm that utilizes reinforcement learning to enforce logical consistency and coherence, allowing models to robustly navigate contradictory information without relying on static heuristics.
\item By leveraging both unstructured text and structured local knowledge graphs, \textsc{Kcr} effectively decomposes complex conflict logic, overcoming the ``Logic Entanglement'' bottleneck.
\item Empirical evaluations demonstrate that \textsc{Kcr} enables 7B-parameter models to outperform SOTA frameworks equipped with top-tier closed-source LLMs (e.g., GPT-5.1), verifying the efficacy of reasoning-driven resolution.
\end{itemize}
\section{Problem Definition}\label{sec:problem}
In the setting of explicit knowledge conflicts, the input typically consists of a query $q$ and a pair of conflicting candidate answers, $A_1$ and $A_2$.
Formally, following the definition in ConflictQA~\cite{benchmark}, each answer is supported by a distinct context containing relevant background and evidence, denoted as $C_1$ and $C_2$, respectively.
The objective is to generate a concise final answer that correctly addresses $q$ by adjudicating the conflict between the provided contexts $(C_1, C_2)$ and their associated logic.

\begin{figure*}[t] 
    \centering
    \includegraphics[width=\textwidth]{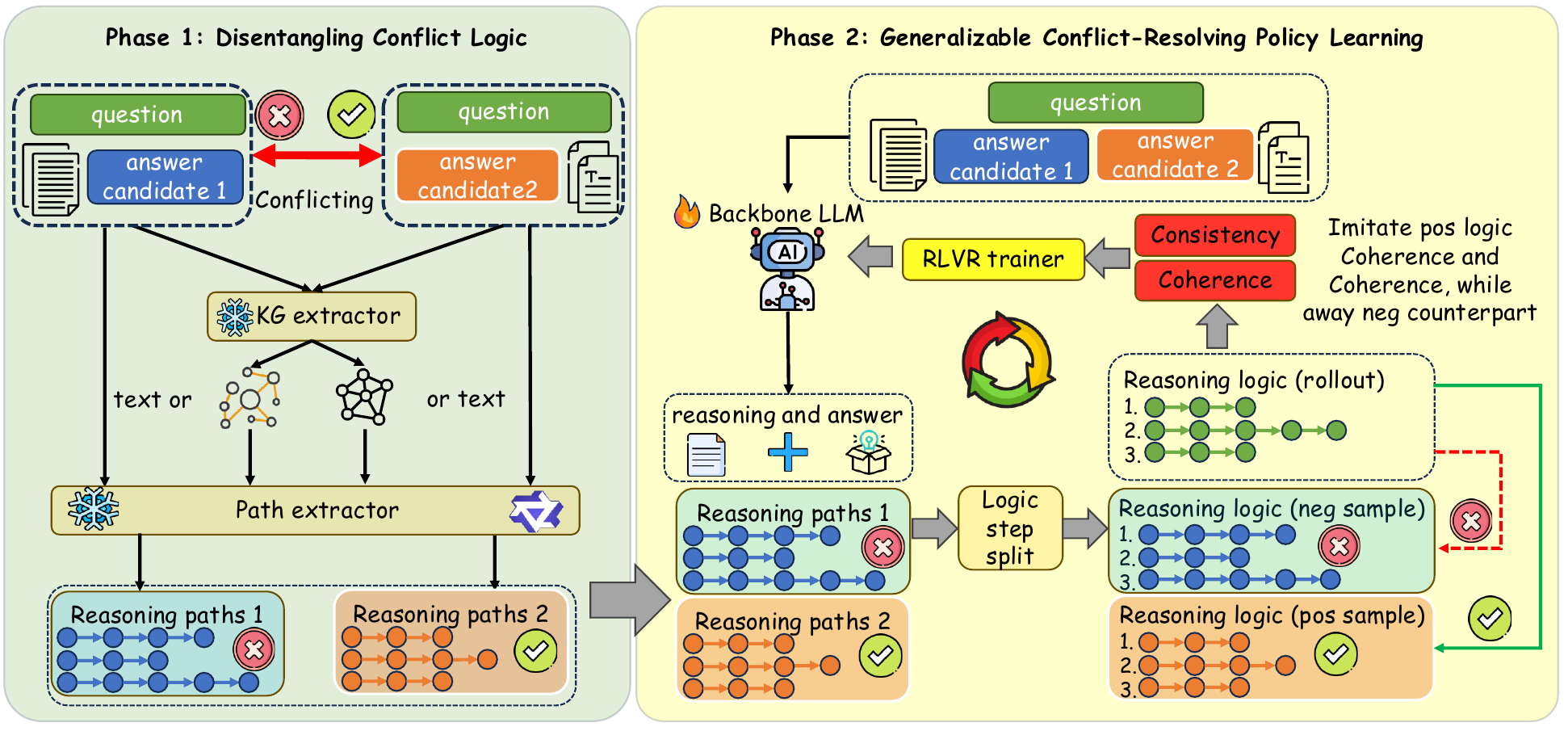} 
    \caption{
    {\textbf{Overview of the \textsc{Kcr} framework.} Our approach structures contradictory evidence into intermediate reasoning traces, followed by a reinforcement learning stage designed to maximize logical consistency and eliminate spurious paths during conflict resolution.}
    } 
    \label{framework}
\end{figure*}

\section{Methodology}
We introduce \textsc{Kcr} (Knowledge Conflict Reasoning), a framework designed to enhance LLM capabilities in resolving explicit knowledge conflicts. As illustrated in Figure~\ref{framework}, \textsc{Kcr} operates in two stages: (i) \textbf{Conflict Logic Disentanglement}, which decomposes entangled contexts into discrete reasoning traces; and (ii) \textbf{Generalizable Policy Learning}, which adapts the backbone model via Reinforcement Learning with Verifiable Rewards (RLVR) to instill a robust conflict-resolution policy.

\subsection{Disentangling Conflict Logic}
This phase aims to extract structured reasoning traces from conflicting contexts. Specifically, \textsc{Kcr} first identifies pivotal entities and relations within the query, then reorganizes and generates reasoning trace sets corresponding to each candidate answer, leveraging both unstructured (text) and structured (graph) data formats.

\subsubsection{Key Entity and Relation Extraction.}
The purpose of obtaining the key entity and relation is to capture the query semantics.
Formally, given the query $q$, \textsc{Kcr} extracts the key entity $e_q$ and relation $r_q$.
For implementation, we utilize open-source entity labels from the popQA dataset and apply a frozen Qwen2.5-7B~\cite{qwen2.5} model for zero-shot Named Entity Recognition and Relation Extraction in other cases. Extraction prompt templates are detailed in Appendix \ref{extraction}.

\subsubsection{Textual Reasoning Traces}
With $e_q$ and $r_q$ extracted, the next step is to disentangle the conflicting context and reorganize the reasoning traces. 
Each trace captures a reasoning path related to $e_q$ or $r_q$. For unstructured textual contexts, we combine answer $A_i$ with its context $C_i$ and prompt a frozen LLM to extract reasoning traces $\mathcal{RT}$ related to the key entities. 
Each trace is defined as a token sequence comprising an alternating series of entities and relations. Formally, the $i$-th reasoning trace from the textual context is defined as follows:
\begin{equation}
    \mathcal{RT}_{i}^{(T)} = (e_1 \to r_1 \to e_2 \to \dots \to e_n),
\end{equation}
where $e_i$ and $r_i$ correspond to entities and relations appearing in the answer and context. We apply this procedure to obtain trace sets $\mathcal{RT}^{(T)}_1$ and $\mathcal{RT}^{(T)}_2$ for answers $A_1$ and $A_2$, respectively.

\subsubsection{Graph Reasoning Traces}
While \textsc{Kcr} can extract reasoning traces from textual form, \textsc{Kcr} also constructs local knowledge graphs to represent context in an interpretable and scalable manner~\cite{GraphRAG}.
Formally, given a query entity $e_q$, relation $r_q$, candidate answer $A_i$, and context $C_i$, we instantiate a local knowledge graph $G_i = (\mathcal{E}_i, \mathcal{R}_i)$. To ensure scalable and interpretable construction, we employ a generator (e.g., GPT-4o-mini) to transform the unstructured context $C_i$ into a structured graph format, where $\mathcal{E}_i$ and $\mathcal{R}_i$ denote the sets of entities and relations, respectively.
For each entity $e \in \mathcal{E}_i$, we define a \textit{graph reasoning trace} ($p_e$) as a directed path sequence:
\begin{equation}
    p_e = e \xrightarrow{r_1} e_2 \xrightarrow{r_2} \cdots \xrightarrow{r_{n-1}} e_n,
\end{equation}
where $e_k \in \mathcal{E}_i$ and $r_k \in \mathcal{R}_i$. 

To isolate reasoning logic relevant to the conflict, we extract the set of traces $\mathcal{RT}^{(G)}_i$ that explicitly intersect with the query semantics. Specifically, we enumerate all paths originating from entities in $\mathcal{E}_i$ that contain either the query entity $e_q$ or the query relation $r_q$:
\begin{equation}
    \mathcal{RT}^{(G)}_i = \{ p_e \mid (e_q, r_q \in p_e) \wedge (e_i \in \mathcal{E}_i)\}.
\end{equation}
We apply this procedure identically to derive trace sets $\mathcal{RT}^{(G)}_1$ and $\mathcal{RT}^{(G)}_2$ corresponding to answer-context pairs $(A_1, C_1)$ and $(A_2, C_2)$.

\subsection{Generalizable Conflict-Resolving Policy Learning}
In this phase, \textsc{Kcr} adapts the backbone LLM to internalize a robust conflict resolution paradigm via reinforcement learning. This design choice is driven by three critical insights regarding the limitations of standard supervision in conflict scenarios.
First, given the \textbf{heterogeneity of conflict cases}, relying solely on Supervised Fine-Tuning (SFT) proves inadequate. SFT often encourages the model to memorize specific conflict patterns rather than learning a generalized adjudication logic capable of handling diverse scenarios.
Second, unlike mathematical or code generation tasks where intermediate reasoning steps are explicitly verifiable, the reasoning process underlying conflict resolution lacks a unique \textbf{``gold standard.''} Step-level supervision is unavailable, as there are multiple valid linguistic paths to adjudicate a conflict. This inherently difficult verification motivates \textsc{Kcr} to optimize for process-level \textit{coherence} rather than token-level accuracy, facilitating inductive learning by contrasting valid reasoning traces against their spurious counterparts.
Finally, the adjudication of contradictory information carries a heightened risk of \textbf{hallucination}, where models may erroneously blend mutually exclusive facts.
This motivates the addition of a verification mechanism to penalize logical inconsistencies that standard likelihood maximization might overlook.

To address these challenges, \textsc{Kcr} introduces two distinct reward signals designed to guide the reinforcement learning process:
\begin{itemize}[leftmargin=*]
    \item \textbf{Logic Coherence Reward:} Incentivizes the LLM to align its generation with the structural logic of correct reasoning traces, while actively diverging from the fallacious logic of incorrect contexts.
    \item \textbf{Logic Consistency Reward:} Penalizes hallucinated or self-contradictory outputs, ensuring that the generated reasoning process and the final answer remain logically consistent with the supported evidence.
\end{itemize}

\subsubsection{Logic Coherence Reward}
The core objective of logical coherence is to guide the backbone LLM to approximate the logical structure present in valid contexts while diverging from it in contradictory ones.

Formally, given a query $q$ and conflicting answers $(A_1, A_2)$ with their associated contexts $(C_1, C_2)$, \textsc{Kcr} prompts the backbone LLM to generate an output containing both a reasoning trace $\mathcal{RT}_0$ and a concise final answer $A_0$. 
\textsc{Kcr} then computes logic coherence scores based on the generated reasoning chain $\mathcal{RT}_0$ relative to the reference reasoning path sets $\mathcal{RT}_1$ and $\mathcal{RT}_2$ extracted in the former phase.
The logic score function $l(\cdot)$ utilizes the Jensen-Shannon (JS) divergence~\cite{JS_Divergence} to measure the semantic continuity between consecutive reasoning steps. For a reasoning chain $\mathcal{R}_i$ consisting of $|\mathcal{R}_i|$ steps:
\begin{equation}\label{eq:logic}
    l(\mathcal{R}_i)= \sum_{j=0}^{len(\mathcal{R}_i)-1} \text{JS}(\mathcal{D}_j||\mathcal{D}_{j+1}),
\end{equation}
where $\text{JS}(\mathcal{D}_j||\mathcal{D}_{j+1})=\frac{1}{2}D_\text{KL}(\mathcal{D}_j||\mathcal{M}) + \frac{1}{2}D_\text{KL}(\mathcal{D}_{j+1}||\mathcal{M})$, with $\mathcal{M} = \frac{1}{2}(\mathcal{D}_j + \mathcal{D}_{j+1})$ and $D_\text{KL}(\cdot)$ denoting the Kullback-Leibler divergence\cite{KL_Divergence}.
The semantic distribution $\mathcal{D}_j$ is derived from the normalized distribution of $i$-th embedding vector $\mathbf{R}_j$ of $\mathcal{R}$,
\begin{equation}
    \mathcal{D}_j = \text{softmax}(\frac{\mathbf{R}_j-\mu(\mathbf{R}_j)}{\sigma(\mathbf{R}_j)}).
\end{equation}
where $\mu(\cdot)$ and $\sigma(\cdot)$ denote the mean and standard deviation of the embedding vector, respectively.

Without loss of generality, let $A_1$ denote the correct (ground-truth) answer and $A_2$ the incorrect counterpart. \textsc{Kcr} defines the discrete \textbf{Logic Coherence Reward} ($\text{R}_{\text{coh}}$) as a binary signal that rewards the model if the logical structure of the generated trace $\mathcal{RT}_0$ aligns closer to the correct reference set $\mathcal{RT}_1$ than the incorrect set $\mathcal{RT}_2$:
\begin{equation}
    \text{R}_{\text{coh}}=
    \begin{cases}
        1, & \text{if } \Delta l(\mathcal{RT}_0, \mathcal{RT}_1) < \Delta l(\mathcal{RT}_0, \mathcal{RT}_2) \\
        0, & \text{otherwise},
    \end{cases}
\end{equation}
where $\Delta l(X, Y) = |l(X) - l(Y)|$. 
Alternatively, a continuous variant of Logic Coherence Reward is: 
\begin{equation}
\hat{\text{R}}_{coh} = \Delta l(\mathcal{RT}_0, \mathcal{RT}_2) -
\Delta l(\mathcal{RT}_0, \mathcal{RT}_1).
\end{equation}
In practice, we find the discrete formulation yields more stable convergence during optimization.

\subsubsection{Logic Consistency Reward}
The Logic Consistency Reward addresses the observation that conflicting contexts often induce hallucinations, where models generate entangled reasoning that mixes contradictory content~\cite{long_context_hallucination}. This reward enforces holistic consistency between the generated reasoning process and the final conclusion.

Formally, \textsc{Kcr} computes the similarity between the generated answer $A_0$ and the candidates $A_1/A_2$, as well as between the generated trace $\mathcal{RT}_0$ and the references $\mathcal{RT}_1/\mathcal{RT}_2$.
The reward is granted if and only if $A_0$ and $\mathcal{RT}_0$ consistently align with the same candidate side, i.e., ($A_1$,$\mathcal{RT}_1$) or ($A_2$,$\mathcal{RT}_2$).
The discrete \textbf{Logic Consistency Reward} ($\text{R}_{\text{con}}$) is defined as:
\begin{equation}
\text{R}_{\text{con}} = 
\begin{cases}
1, & \text{if } (S_{RT}^{0,1} > S_{RT}^{0,2}) \land (S_{A}^{0,1} > S_{A}^{0,2}) \\
1, & \text{if } (S_{RT}^{0,2} > S_{RT}^{0,1}) \land (S_{A}^{0,2} > S_{A}^{0,1}) \\
0, & \text{otherwise},
\end{cases}
\end{equation}
where $S_{RT}^{i,j} = S(\mathcal{RT}_i, \mathcal{RT}_j)$ and $S_{A}^{i,j} = S(A_i, A_j)$.  $S(\cdot, \cdot)$ denotes the normalized Levenshtein similarity~\cite{Levenshtein_distance}, with higher scores indicating greater string similarity, chosen for its computational efficiency at the token level.
Alternatively, we formulate a continuous variant of the Logic Consistency Reward as a margin-based metric. This formulation quantifies the magnitude of the alignment differential between competing reasoning traces and candidate answers:
\begin{align}
\hat{\text{R}}_{\text{con}} = \ & \left| S_{RT}^{0,1} - S_{RT}^{0,2} \right| + \left| S_{A}^{0,1} - S_{A}^{0,2} \right|,
\end{align}
where $S_{RT}^{0,i}$ and $S_{A}^{0,i}$ denote the similarity scores of the generated output against the $i$-th candidate reasoning trace and answer, respectively. 

Note that ground-truth labels are not utilized for the consistency reward. This prevents the model from overfitting to correct answers while ignoring flawed reasoning traces, ensuring true process-outcome alignment.
Token-level similarity also helps reduce computational overhead.

\subsubsection{RLVR Training Procedure}
After deriving the reward signals, \textsc{Kcr} optimizes the backbone LLM as a policy $\pi_\theta$ using Group Relative Policy Optimization (GRPO)~\cite{GRPO}. We select discrete variants for both Logic Coherence and Logic Consistency rewards. This design choice is critical for batch-wise RLVR methods, as discrete signals significantly enhance discriminability during group sampling and stabilize candidate selection compared to continuous values. Furthermore, we enforce strict format constraints to ensure structural validity while maintaining the original ground-truth correctness signal.

Formally, for each query $q$, we sample a group of outputs $\{o_1, \dots, o_G\}$ from the reference policy $\pi_{\theta_{\text{old}}}$. The cumulative reward for the $i$-th output is aggregated as $r_i = \text{R}_{\text{coh}} + \text{R}_{\text{con}} + \text{R}_{\text{correct}}$. To stabilize optimization, we compute the advantage score $\hat{A}_{i,t}$ by normalizing rewards within the group:
\begin{equation}
    \hat{A}_{i,t} = \frac{r_i - \mu(\{r_j\}_{j=1}^G)}{\sigma(\{r_j\}_{j=1}^G) + \epsilon},
\end{equation}
where $\mu$ and $\sigma$ denote the group mean and standard deviation, respectively, and $\delta$ is a small constant for numerical stability. The policy ratio $r_{i,t}(\theta)$ between the current and reference policy is defined as:
\begin{equation}
    r_{i,t}(\theta) = \frac{\pi_{\theta}(o_{i,t} \mid q, o_{i,<t})}{\pi_{\theta_{\text{old}}}(o_{i,t} \mid q, o_{i,<t})},
\end{equation}
where $o$ is the output and $\pi$ is the backbone LLM as policy model and $\theta$ the weight of backbone LLM.
The final GRPO objective maximizes the surrogate loss while penalizing significant deviations from the reference policy via clipping:
\begin{equation}
\begin{split}
    \mathcal{J}(\theta) &= \mathbb{E}_{(q, A) \sim \mathcal{D},\ \{o_i\}_{i=1}^G \sim \pi_{\theta_{\text{old}}}} \Bigg[ \frac{1}{G} \sum_{i=1}^{G} \frac{1}{|o_i|} \sum_{t=1}^{|o_i|} \\
    & \min \Big( r_{i,t}(\theta) \hat{A}_{i,t}, \text{clip}(r_{i,t}(\theta), \varepsilon) \hat{A}_{i,t} \Big) \Bigg],
\end{split}
\end{equation}
where \text{clip}($\cdot$, $\varepsilon$ ) is the standard clip operation with $\varepsilon$ as the clipping hyperparameter.

\section{Experiments}
\label{sec:experiments}

We evaluate \textsc{Kcr} through quantitative and qualitative experiments. First, we validate its effectiveness across various backbone LLMs in scenarios involving conflicting knowledge in long contexts. We then perform ablation studies to assess the logic coherence and consistency rewards, comparing \textsc{Kcr} against Supervised Fine-Tuning (SFT) and standard GRPO. Finally, case analyses demonstrate the improved reasoning capabilities.

\subsection{Experimental Setting}

\noindent \textbf{Datasets.}
We utilize two public datasets specifically adapted for the Explicit Knowledge Conflict task: \textbf{popQA} and \textbf{strategyQA} \cite{benchmark}. 
PopQA involves conflicts between distinct entities, whereas strategyQA features general interrogatives with conflicting Boolean answers. 
We partition data into training, validation, and test sets with an 8:1:1 ratio.
Table~\ref{tab:statistics} summarizes the statistics.

\begin{table}[htp]
\footnotesize
\centering
    \begin{tabular}{lrr}
    \toprule
    Dataset & popQA & strategyQA \\ \midrule
    Questions & 7,198 & 1,244 \\ 
    Avg. question tokens  & 11.73  & 12.09 \\
    Avg. answer tokens    & 23.47  & 1.00 \\
    Avg. context tokens   & 498.65 & 397.69 \\
    Relative token ratio  & 21.24  & 397.69 \\
    \bottomrule
    \end{tabular}
    \caption{Dataset statistics. Relative token ratio is the number of context tokens divided by answer tokens.}
    \label{tab:statistics}
\end{table}

\noindent \textbf{Hyperparameter Settings and Training Cost.}
We employ consistent hyperparameters across experiments, training for 10 epochs with a batch size of 32. 
GPT-4o-mini powers both local knowledge graph construction and the LLM-as-Judge module, while \texttt{BAAI/bge-large-en-v1.5} generates sentence embeddings.
Training is accelerated using the Verl framework with Flash Attention. All models are trained on 8$\times$A100 80GB SXM GPUs. The approximate training time is 72 GPU hours for 7B/8B models and 25 GPU hours for 3B models.

\noindent \textbf{Evaluation Metrics.}
We adopt both semantic-level and token-level evaluations.
Specifically, For semantic similarity ($\text{ACC}_L$), we use an LLM-as-Judge \cite{LLM-as-a-Judge} powered by GPT-4o-mini (prompts in Appendix). For token-level accuracy, we report Exact Match ($\text{ACC}_{EM}$) and Cover Exact Match ($\text{ACC}_{CEM}$).
The judge prompts are detailed in the {Appendix \ref{judge_prompt}}.

\noindent \textbf{Baselines.}
We evaluate \textsc{Kcr} using LLaMA and Qwen backbones.
We compare against five state-of-the-art (SOTA) frameworks for inter-context knowledge conflicts, including four decoding strategies: CAD \cite{CAD}, CUAD \cite{CUAD}, AdaCAD \cite{AdaCAD}, and CoCoA \cite{CoCoA}. Additionally, we include Astute RAG \cite{Astute_RAG}, which employs explicit knowledge fusion.

\begin{table*}[t!]
    \centering
    \footnotesize
    \renewcommand{\arraystretch}{0.9}
    \begin{tabular}{llcccccc}
        \toprule
        & & \multicolumn{3}{c}{\textbf{popQA}} & \multicolumn{3}{c}{\textbf{strategyQA}} \\
        \cmidrule(lr){3-5} \cmidrule(lr){6-8}
        \textbf{Model Family} & \textbf{Method / Variant} & $\text{ACC}_L$ & $\text{ACC}_{EM}$ & $\text{ACC}_{CEM}$ & $\text{ACC}_L$ & $\text{ACC}_{EM}$ & $\text{ACC}_{CEM}$ \\
        \midrule

        \multicolumn{2}{l}{\textit{\textbf{Direct Generation (Without Context)}}} \\
        & Llama 3.2-3B & 0.1764 & 0.0056 & 0.1736 & 0.4640 & 0.0000 & 0.3200 \\
        & Qwen 2.5-7B & 0.1250 & 0.0889 & 0.0944 & 0.6880 & 0.6720 & 0.6720 \\
        & Qwen 3-8B (Thinking) & 0.1736 & 0.1306 & 0.1444 & 0.7600 & 0.7200 & 0.7680 \\
        & GPT-4o-mini & 0.3042 & 0.2431 & 0.2542 & 0.7440 & 0.7440 & 0.7440 \\
        & GPT-5.1 (Thinking) & 0.4944 & 0.1389 & 0.4819 & 0.6640 & 0.0000 & 0.6720 \\
        \midrule
        \multicolumn{2}{l}{\textit{\textbf{General Baselines (With Context)}}} \\
        & Llama 3.2-1B & 0.3000 & 0.0139 & 0.2625 & 0.4480 & 0.0160 & 0.0640 \\
        & Llama 3.1-8B & 0.5653 & 0.0292 & 0.5278 & 0.4480 & 0.0000 & 0.4000 \\
        & Qwen 2.5-1.5B & 0.3986 & 0.2569 & 0.3764 & 0.5040 & 0.3840 & 0.3920 \\
        & Qwen 2.5-3B & 0.5167 & 0.4333 & 0.5042 & 0.5440 & 0.5360 & 0.5360 \\
        & \textsc{CUAD} (Default) & 0.3431 & 0.2958 & 0.3056 & 0.5040 & 0.5040 & 0.5040 \\
        & \textsc{AdaCAD} (Default) & 0.3431 & 0.2958 & 0.3056 & 0.5040 & 0.5040 & 0.5040 \\
        & \textsc{CoCoA} (Default) & 0.3431 & 0.2958 & 0.3056 & 0.5040 & 0.5040 & 0.5040 \\
        & \textsc{Astute RAG} (GPT-4o) & 0.5486 & 0.5056 & 0.6097 & 0.5920 & 0.5760 & 0.5840 \\
        & \textsc{Astute RAG} (GPT-5.1) & 0.6694 & 0.5819 & 0.7083 & 0.5760 & 0.2160 & 0.5920 \\
        \midrule

        \multicolumn{2}{l}{\textit{\textbf{Backbone: Llama 3.2-3B}}} \\
        & Directly Use Backbone & 0.4958 & 0.3375 & 0.4944 & 0.5120 & 0.3920 & 0.4240 \\
        & \textsc{CAD} Decoding & 0.2972 & 0.3111 & 0.3222 & 0.4640 & 0.3280 & 0.4880 \\
        & \textsc{Astute RAG} & 0.3972 & 0.0208 & 0.7458 & 0.4640 & 0.3840 & 0.4320 \\
        \rowcolor[gray]{0.96} & \textbf{KCR} (``Text'') & 0.8417 & 0.8014 & 0.8292 & 0.7760 & 0.7680 & 0.7680 \\
        \rowcolor[gray]{0.96} & \textbf{KCR} (``Graph'') & \textbf{0.8750} & \textbf{0.8361} & \textbf{0.8778} & \textbf{0.8240} & \textbf{0.7760} & \textbf{0.7760} \\
        & \textit{Improvement vs Backbone} & \textit{+37.9\%} & \textit{+49.9\%} & \textit{+38.3\%} & \textit{+31.2\%} & \textit{+38.4\%} & \textit{+35.2\%} \\
        \midrule

        \multicolumn{2}{l}{\textit{\textbf{Backbone: Qwen 2.5-7B}}} \\
        & Directly Use Backbone & 0.4847 & 0.4181 & 0.4778 & 0.5920 & 0.5920 & 0.5920 \\
        & \textsc{CAD} Decoding & 0.4028 & 0.1750 & 0.4361 & 0.3120 & 0.0000 & 0.3440 \\
        & \textsc{Astute RAG} & 0.4375 & 0.3889 & 0.4556 & 0.3680 & 0.3600 & 0.3600 \\
        \rowcolor[gray]{0.96} & \textbf{KCR} (``Text'') & \textbf{0.8667} & \textbf{0.8250} & \textbf{0.8653} & \textbf{0.8320} & \textbf{0.7920} & \textbf{0.7920} \\
        \rowcolor[gray]{0.96} & \textbf{KCR} (``Graph'') & 0.8014 & 0.7569 & 0.7778 & 0.7520 & 0.6720 & 0.6720 \\
        & \textit{Improvement vs Backbone} & \textit{+38.2\%} & \textit{+40.7\%} & \textit{+38.8\%} & \textit{+24.0\%} & \textit{+20.0\%} & \textit{+20.0\%} \\
        \midrule

        \multicolumn{2}{l}{\textit{\textbf{Backbone: Qwen 3-8B (Thinking)}}} \\
        & Directly Use Backbone & 0.5431 & 0.5056 & 0.5472 & 0.5600 & 0.5600 & 0.5600 \\
        \rowcolor[gray]{0.96} & \textbf{KCR} (``Text'') & 0.6861 & 0.6458 & 0.6583 & \textbf{0.6400} & \textbf{0.6400} & \textbf{0.6400} \\
        \rowcolor[gray]{0.96} & \textbf{KCR} (``Graph'') & \textbf{0.6958} & \textbf{0.6556} & \textbf{0.6903} & 0.5600 & 0.5600 & 0.5600 \\
        & \textit{Improvement vs Backbone} & \textit{+15.3\%} & \textit{+15.0\%} & \textit{+14.3\%} & \textit{+8.0\%} & \textit{+8.0\%} & \textit{+8.0\%} \\
        \midrule

        \multicolumn{2}{l}{\textit{\textbf{Backbone: Qwen 3-8B}}} \\
        & Directly Use Backbone & 0.5431 & 0.5056 & 0.5500 & 0.5360 & 0.5360 & 0.5360 \\
        & \textsc{CAD} Decoding & 0.4028 & 0.1750 & 0.4361 & 0.3120 & 0.0000 & 0.3440 \\
        & \textsc{Astute RAG} & 0.5028 & 0.4417 & 0.6764 & 0.5040 & 0.4880 & 0.7040 \\
        \rowcolor[gray]{0.96} & \textbf{KCR} (``Text'') & 0.7472 & 0.7139 & 0.7361 & 0.5861 & 0.5444 & 0.5806 \\
        \rowcolor[gray]{0.96} & \textbf{KCR} (``Graph'') & \textbf{0.9028} & \textbf{0.8764} & \textbf{0.9000} & 0.5278 & 0.4889 & 0.5319 \\
        & \textit{Improvement vs Backbone} & \textit{+36.0\%} & \textit{+37.1\%} & \textit{+35.0\%} & \textit{+5.0\%} & \textit{+0.8\%} & \textit{+4.5\%} \\

        \midrule
        \multicolumn{2}{l}{\textbf{Overall Improvement vs SOTA}} & \textbf{+19.7\%} & \textbf{+24.3\%} & \textbf{+12.0\%} & \textbf{+7.2\%} & \textbf{+4.8\%} & \textbf{+2.4\%} \\
        
        \bottomrule
    \end{tabular}
    \caption{\textbf{Overall comparison on the Inter-Context Knowledge Conflict task.} \textbf{Bold} indicates the best performance within each specific backbone group or overall. ``\textsc{Text}'' and ``\textsc{Graph}'' denote \textsc{Kcr} reasoning traces derived from textual and graph-based formats, respectively. The improvement rows highlight the relative gain of the best \textsc{Kcr} variant over its corresponding backbone baseline, while the final row compares \textsc{Kcr} against the strongest state-of-the-art framework. All our experimental results are averaged over three independent runs.}
    \label{results_for_all}
\end{table*}

\subsection{Overall Performance Comparison}
The main quantitative results for the Explicit Knowledge Conflict task are presented in Table~\ref{results_for_all}, from which several key observations can be drawn.
First, \textbf{KCR} consistently and significantly improves the performance of backbone LLMs across datasets of varying scales. This is particularly evident in $\text{ACC}_L$, which emphasizes semantic-level answer accuracy, suggesting \textsc{Kcr} enhances true understanding rather than mere lexical matching.
Notably, both text-based and graph-based reasoning paths contribute to these improvements. This indicates that \textsc{Kcr} effectively guides backbone LLMs in conflict adjudication, even when the models have not been explicitly pre-trained on graph-structured inputs.
While standard decoder-based baselines struggle to resolve explicit knowledge conflicts effectively, \textsc{Kcr} demonstrates universal performance gains. Additionally, high-complexity frameworks like Astute RAG often fail on smaller (3B- and 7B-scale) models due to excessive reasoning demands. In contrast, \textsc{Kcr} successfully instills the conflict-resolution paradigm into smaller backbones, enabling a 7B model to outperform much larger, closed-source systems.

\begin{figure*}[htb]
\centering
\includegraphics[width=\textwidth]{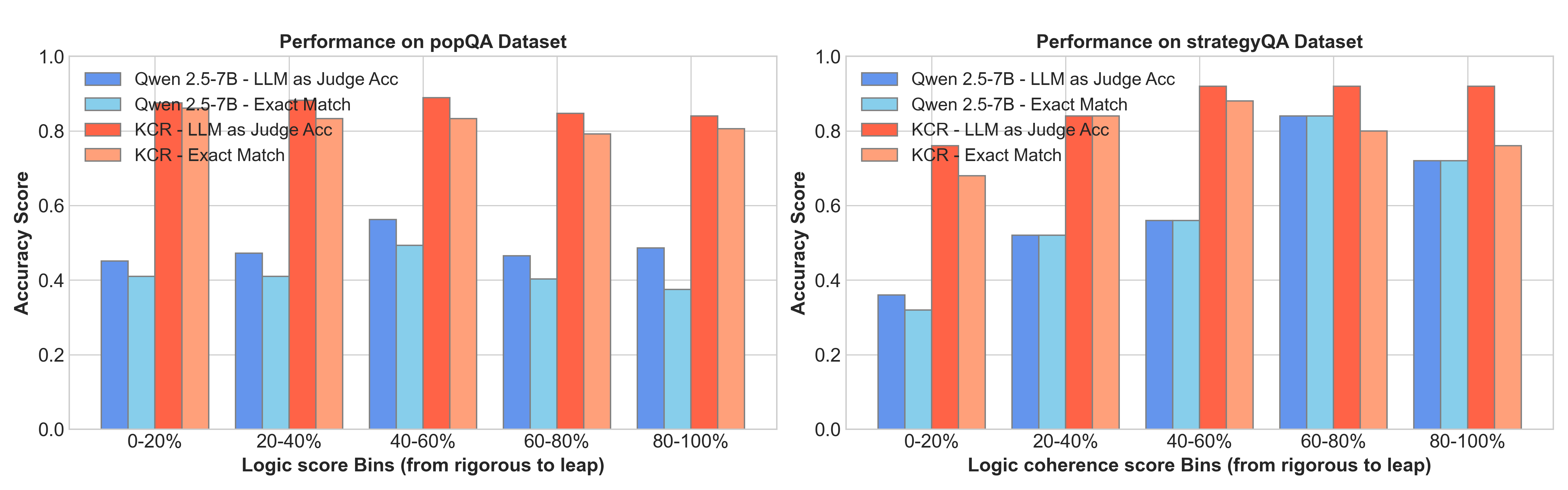}
\caption{{\textbf{Impact of Reasoning Coherence.}} Analysis of model performance across varying degrees of logical coherence (JS Divergence). Lower scores indicate stricter logical adherence.}
\label{logic}
\end{figure*}

\begin{figure}[htb]
\centering
\includegraphics[width=\linewidth]{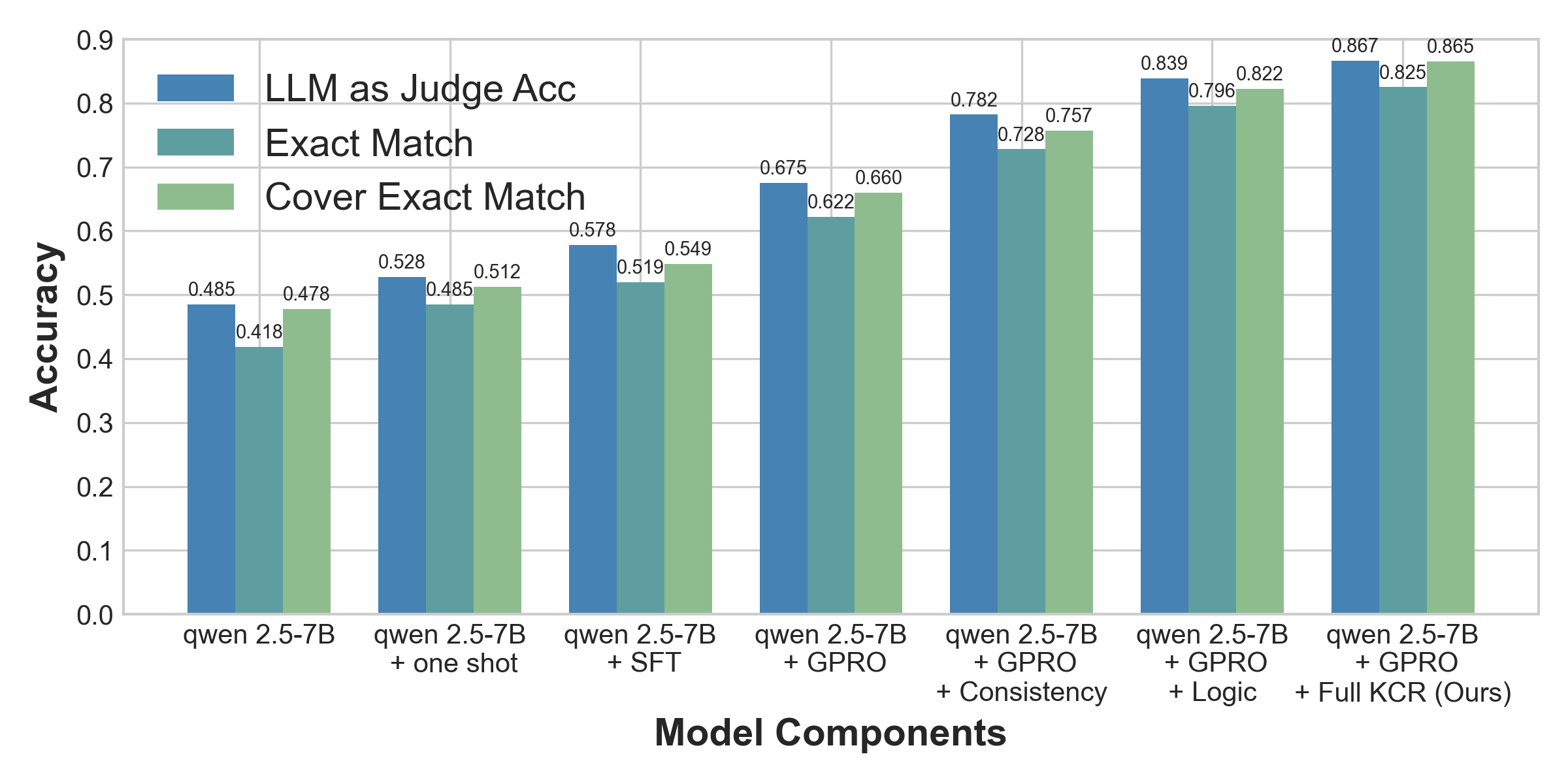}
\caption{{\textbf{Ablation Study.}} \textsc{Kcr} excels in scenarios requiring rigorous logical chains (popQA) compared to looser reasoning tasks. 
        (b) validates that both Logic Coherence and Consistency rewards are essential for robust conflict adjudication.}
\label{ablation_study}
\end{figure}

\subsection{Impact of Logical Coherence}
In \textsc{Kcr}, the logical coherence of the reasoning process serves as a proxy for rigor, where lower coherence scores (in JS divergence) indicate tighter logical steps, while higher scores reflect larger inferential leaps. 
It is worth noting that questions requiring excessive rigor or extreme logical leaps are inherently more difficult.
As shown in Figure~\ref{logic}, \textsc{Kcr} notably improves the handling of conflicting answers on the more challenging popQA dataset. On the simpler StrategyQA, \textsc{Kcr} improves performance specifically on questions that demand rigorous logic. This suggests that \textsc{Kcr} is particularly advantageous for problems requiring complex, multi-step adjudication.

\subsection{Ablation Study}
We evaluate the individual contributions of \textsc{Kcr}’s components on popQA using the Qwen 2.5–7B backbone.
This section evaluates \textsc{Kcr}’s components on popQA and compares its performance to alternative fine-tuning approaches using the Qwen 2.5–7B backbone.
Figure~\ref{ablation_study} illustrates that while each component contributes to the model's overall capability, the \textbf{Logic Coherence Reward} is the principal driver of improvement. The full \textsc{Kcr} framework effectively integrates these elements to form a robust system for resolving knowledge conflicts, significantly outperforming approaches that rely solely on SFT or vanilla reinforcement learning methods like GRPO.

\subsection{Qualitative Study}
Our qualitative analysis indicates that models enhanced by 
{\textsc{Kcr} exhibit a remarkably structured reasoning process. They systematically analyze conflicting points sequentially, strictly adhering to the order of evidence presentation before deriving a final conclusion.
In contrast, baseline models without \textsc{Kcr} tend to generate conclusions more directly, often failing to articulate a clear, step-by-step logical chain.} This observable difference in reasoning style demonstrates that \textsc{Kcr} instills a more methodical and logically coherent adjudication process in the backbone model. Detailed case studies, including prompts and outputs, are provided in the {Appendix \ref{case_study}.}
\section{Related Work}
\label{sec:related_work}

\subsection{Implicit Knowledge Conflicts}
Implicit Knowledge Conflicts, including both Context-Memory and Inter-Memory conflicts, arise from discrepancies between externally context (i.e., retrieved documents) and the model's parameterized knowledge, or among the model's parameterized knowledge~\cite{conflict_survey}.
The typical scenario is in Retrieval-Augmented Generation (RAG) frameworks, where the model must reconcile retrieved documents with its own stored knowledge~\cite{RAG_conflict_hallucinations,Rag_solution_1,Rag_solution_2}.
Prevailing strategies for mitigating such conflicts include model-centric approaches like fine-tuning~\cite{FT-1,FT-2} and knowledge editing~\cite{knowledge_inject}, alongside context-centric methods such as information compression and fusion~\cite{Rag_solution_1,Rag_solution_2}.
Implicit Knowledge Conflicts fundamentally do not require models to understand complex conflict logic, and make choices within internalized knowledge and context.

\subsection{Explicit Knowledge Conflicts}
Explicit Knowledge Conflicts typically refer to contradictions that arise within the provided context, most commonly when two conflicting pieces of information are presented as separate contextual inputs~\cite{conflict_survey}.
Typical methods in resolving Explicit Knowledge Conflicts involve re-pairing the conflicting contexts at the sentence level, obtaining soft labels through supervised signals, and then feeding them into the FiD decoder to generate an answer~\cite{FiD, merging_1}.
More recent methods shifted to semantic distribution-level fusion, process each conflicting source document independently to generate separate answers and semantic distributions, which are then fused by a specialized decoder to synthesize a unified output~\cite{CAD, CUAD, AdaCAD, CoCoA}.
Recent approaches attempt to directly perform knowledge fusion using powerful closed-source LLMs, employing complex iterative heuristic frameworks to address this challenge~\cite{Astute_RAG}.
These modeling attempts have not reduced the difficulty of understanding conflicting contexts, making it still challenging to fully leverage LLMs' reasoning abilities to resolve explicit knowledge conflicts.

\subsection{LLMs Behaviors in Knowledge Conflicts}
Recent studies confirm that LLMs possess a baseline capacity for resolving simple knowledge conflicts~\cite{benchmark}. However, these investigations have also uncovered significant behavioral biases. For instance, LLMs tend to favor information characterized by high verbosity or apparent evidential support, and they exhibit difficulty reasoning over ``counter-memory'' facts even when paired with supporting context~\cite{LLM_behave_1, LLM_behave_2, LLM_fix_conflict_1}. Crucially, prior research has concentrated on characterizing these observable behaviors, while largely neglecting the role of the model's underlying logical reasoning abilities in systematically resolving such contradictions.
\section{Conclusion}
\label{sec:conclusion}

We introduce \textsc{Kcr}, a framework designed to resolve explicit knowledge conflicts by distilling structured reasoning paths from contradictory evidence. By optimizing for logical coherence and consistency through a novel learning paradigm, \textsc{Kcr} enables models to systematically adjudicate conflicting information. Empirical results across diverse backbones demonstrate that \textsc{Kcr} significantly improves a model's capacity to navigate factual contradictions, yielding substantial and robust performance gains.

\section*{Acknowledgments}
This research is  partly supported by the Shenzhen Science and Technology Program No. SYSPG20241211173609009, and National Science and Technology Council (NSTC), Taiwan (Grants: NSTC-114-2222-E-A49-004, and NSTC-114-2639-E-A49-001-ASP).

\section*{Limitations}
While \textsc{Kcr} demonstrates significant improvements in resolving explicit knowledge conflicts, we identify several limitations that merit further investigation.

\paragraph{Computational Complexity and Latency.}
The \textit{Conflict Logic Disentanglement} phase relies on structured context via Local Knowledge Graphs (LKGs). Our current implementation utilizes proprietary models (e.g., GPT-4o-mini) for graph extraction, which introduces a dependency on external APIs and increases latency compared to purely embedding-based retrieval. Furthermore, the RLVR training paradigm requires sampling multiple reasoning chains (group sampling) to compute coherence and consistency rewards. This incurs a higher computational cost during training relative to standard Supervised Fine-Tuning (SFT).

\paragraph{Modeling Scope of Conflicts.}
Our framework is explicitly designed to address inter-context conflicts (Explicit Knowledge Conflicts), where retrieved documents contain contradictory information. We do not explore implicit conflicts (Context-Parametric Memory conflicts), where retrieved evidence contradicts the model’s internal knowledge. Additionally, our task formulation focuses on dyadic conflict resolution (binary adjudication). Real-world scenarios involving multi-party conflicts or sources with varying degrees of partial correctness may require more complex adjudication logic than our current implementation.

\paragraph{Linguistic and Benchmark Scope.}
Our experiments and prompt designs are restricted to English benchmarks. The efficacy of the logic disentanglement phase and the graph extraction prompts in multilingual settings or for low-resource languages remains an open question. Future work is required to assess the cross-lingual generalizability of structured reasoning in conflict resolution.

\paragraph{Reward and Noisy.}
Verifiable rewards are less reliable in open-domain or subjective conflicts without clear ground truth, which may introduce ambiguity in learning. Our evaluation is limited to controlled settings, and extending KCR to adversarial or noisy environments is an important direction for future work.

\bibliography{ref}
\newpage
\appendix
\section*{Appendix}
\label{sec:appendix}

\lstset{
  basicstyle=\ttfamily\footnotesize,
  breaklines=true,
  frame=none,
  columns=fullflexible
}

\section{Checklist}
\subsection{Ethics Statement}\label{Ethics}
All datasets used in this research are publicly available and were sourced from previous studies that have undergone appropriate ethical review. Our work did not involve the collection of any new data from human subjects. We have adhered to all data usage agreements and licenses associated with these pre-existing datasets.

\subsection{Reprodicibility Statement}\label{Reprodicibility_statement}
We are committed to making our research reproducible. All datasets used in this study are publicly available, and we provide detailed descriptions and sources in our experimental setup section. 

\subsection{Potential Risks}\label{Risks}
We do not identify significant risks associated with our approach, as it relies solely on open-source data and models and follows standard research practices.

\subsection{Usage of LLM}\label{LLM_usage}

This paper primarily employs large language models (LLMs) to refine overall writing quality, with a particular focus on eliminating incorrect expressions, minimizing grammatical errors, and enhancing clarity, coherence, and readability to ensure the text meets ACL standards.

\section{Prompt Template}

\subsection{Reasoning and Answer Extraction}
Extending \textsc{Kcr} to new backbone models primarily involves defining regex patterns to isolate intermediate reasoning traces from the final answer. The patterns currently employed for the Qwen and LLaMA series are provided in the boxes below.

\begin{tcolorbox}[
    colback=gray!5, 
    colframe=gray!60!black,
    title=Patterns for extracting thinking and answer from Qwen,
    fonttitle=\small\sffamily,
    boxrule=0.8pt,
    sharp corners,
    left=1mm, right=1mm, top=1mm, bottom=1mm,
    breakable
]
\begin{lstlisting}[
    breaklines=true,
    columns=fullflexible,
    keepspaces=true,
    breakatwhitespace=false,
    xleftmargin=0pt,
    frame=none
]
thinking_patterns = r"<think>(.*?)</think>"

answer_patterns = r"<answer>(.*?)</answer>"
\end{lstlisting}
\end{tcolorbox}

\begin{tcolorbox}[
    colback=gray!5, 
    colframe=gray!60!black,
    title=Patterns for extracting thinking and answers from LLaMA,
    fonttitle=\small\sffamily,
    boxrule=0.8pt,
    sharp corners,
    left=1mm, right=1mm, top=1mm, bottom=1mm,
    breakable
]
\begin{lstlisting}[
    breaklines=true,
    columns=fullflexible,
    keepspaces=true,
    breakatwhitespace=false,
    xleftmargin=0pt,
    frame=none
]
thinking patterns = r'(?:\*+)?\s*Thinking Process\s*:?\s*(?:\*+)?\s*(.*?)\s*(?:\*+)?\s*Final Answer\s*:?\s*(?:\*+)?'

answer_patterns = r"\*\*Final Answer:\*\*\s*(.+)"
answer_patterns = r"Final Answer:\s*(.+)"
answer_patterns = r"\*\*Correct Answer:\*\*\s*(.+)"
answer_patterns = r"Correct Answer:\s*(.+)"

\end{lstlisting}
\end{tcolorbox}

\subsection{Prompts for KG Extraction}
\label{extraction}
In the KG extraction phase, we utilize GPT-4o-mini as the extractor. The system and user prompts are structured to ensure the output adheres to a strict JSON format containing entities and triples.

\begin{tcolorbox}[
    colback=gray!5, 
    colframe=gray!60!black,
    title=Prompts for extracting thinking and answers using GPT-4o-mini,
    fonttitle=\small\sffamily,
    boxrule=0.8pt,
    sharp corners,
    left=1mm, right=1mm, top=1mm, bottom=1mm,
    breakable
]
\begin{lstlisting}[
    breaklines=true,
    columns=fullflexible,
    keepspaces=true,
    breakatwhitespace=false,
    xleftmargin=0pt,
    frame=none
]
{"role": "system", "content": "You are an expert agent specialized in build Knowledge Graphs."},
{"role": "system", "content": "Extract a knowledge graph from the following document, return a json file within one line."},
{"role": "user", "content": """You must generate the output in a JSON containing a list with JSON objects having the following keys: "entities", "triples". The "entities" must contain the text of the extracted entities from document, the "triples" must contain the python dicts that composed of key "subject", key "relation" and key "object". in this dict."""},
{"role": "user", "content": f"document: {document}"},
{"role": "user", "content": f"key subject: {subject}\n\n"},
{"role": "user", "content": f"key relation: {relation}\n\n"},
{"role": "user", "content": "knowledge graph: \n"},

\end{lstlisting}
\end{tcolorbox}

\subsection{Prompts for Qwen Series Backbone}
The following prompts guide the backbone models to produce outputs following a structured ``Reasoning + Final Answer'' format. This allows the model to explicitly disentangle conflicting evidence before making a decision.

\begin{tcolorbox}[
    colback=gray!5, 
    colframe=gray!60!black,
    title=Prompts for extracting thinking and answers from Qwen,
    fonttitle=\small\sffamily,
    boxrule=0.8pt,
    sharp corners,
    left=1mm, right=1mm, top=1mm, bottom=1mm,
    breakable
]
\begin{lstlisting}[
    breaklines=true,
    columns=fullflexible,
    keepspaces=true,
    breakatwhitespace=false,
    xleftmargin=0pt,
    frame=none
]
{"role": "system", "content": "You are helpful AI system.\n\n"},
{"role": "user", "content": "A question is given and its two candidate answers, along with their context. Only one of the two is correct. You need to choose the correct one. Use English Only!\n\n"},
{"role": "user", "content": f"question: {question} \n"},
{"role": "user", "content": f"First answer: {answer1} \n\n"},
{"role": "user", "content": f"Context with first answer: {context1} \n\n"},
{"role": "user", "content": f"Second answer: {answer2} \n\n"},
{"role": "user", "content": f"Context with second answer: {context2}\n\n"},
{"role": "user", "content": "First output the thinking process in <think> </think> and final answer using single entity in <answer> </answer> tags."},

\end{lstlisting}
\end{tcolorbox}

\subsection{Prompts for LLaMA Series Backbone}
This subsection presents the prompt design for the LLaMA backbone. The following prompts guide the backbone models to produce outputs following a structured ``Reasoning + Final Answer'' format. This allows the model to explicitly disentangle conflicting evidence before making a decision.

\begin{tcolorbox}[
    colback=gray!5, 
    colframe=gray!60!black,
    title=Prompts for extracting thinking and answers from LLaMA,
    fonttitle=\small\sffamily,
    boxrule=0.8pt,
    sharp corners,
    left=1mm, right=1mm, top=1mm, bottom=1mm,
    breakable
]
\begin{lstlisting}[
    breaklines=true,
    columns=fullflexible,
    keepspaces=true,
    breakatwhitespace=false,
    xleftmargin=0pt,
    frame=none
]
{"role": "system", "content": "You are helpful AI system.\n\n"},
{"role": "user", "content": "A question is given and its two candidate answers, along with their context and evidence. Only one of the two is correct. You need to choose the correct one."},
{"role": "user", "content": "You have four key principles that should be followed: a) Check for explicit contradiction or mutual exclusivity; b) Analyze implications or presuppositions; c) Apply world knowledge to check commonsense facts; d) Use context to ground both statements. \n\n"},
{"role": "user", "content": f"Question: {question} \n"},
{"role": "user", "content": f"First answer: {answer1} \n\n"},
{"role": "user", "content": f"Context with first answer: {memory1} \n\n"},
{"role": "user", "content": f"Evidence with first answer: {evidence1} \n\n"},
{"role": "user", "content": f"Second answer: {answer2} \n\n"},
{"role": "user", "content": f"Context with second answer: {memory2} \n\n"},
{"role": "user", "content": f"Evidence with second answer: {evidence2} \n\n"},
{"role": "user", "content": "Firstly, output the whole thinking process in <think> </think> and final answer using single entity in <answer> </answer> tags. Use English to answer the question only!\n\n"},
{"role": "user", "content": "Thinking process and Final answer: "},
\end{lstlisting}
\end{tcolorbox}

\subsection{Prompts for LLM-as-Judge}
To evaluate model performance, we use a judge prompt that verifies the semantic alignment between the predicted answer and the ground truth reference.

\label{judge_prompt}
\begin{tcolorbox}[
    colback=gray!5, 
    colframe=gray!60!black,
    title=LLM-as-judge prompt,
    fonttitle=\small\sffamily,
    boxrule=0.8pt,
    sharp corners,
    left=1mm, right=1mm, top=1mm, bottom=1mm,
    breakable
]
\begin{lstlisting}[
    breaklines=true,
    columns=fullflexible,
    keepspaces=true,
    breakatwhitespace=false,
    xleftmargin=0pt,
    frame=none
]
prompt = '''Given a Question and its Golden Answer, verify whether the Predicted Answer is correct. The prediction is correct if it fully aligns with the meaning and key information of the Golden Answer. Respond with True if the prediction is correct and False otherwise.

    Question: {question}
    
    Golden Answer: {reference}
    
    Predicted Answer: {prediction}
    
    '''
\end{lstlisting}
\end{tcolorbox}

\section{Extended Qualitative Studies}\label{case_study}
In this section, we present some qualitative detail results on the popQA dataset. We compare our method with the backbone model Qwen 2.5-7B.
Here are some cases that demonstrate improved reasoning skills for resolving conflicts in knowledge using \textsc{Kcr}.

\begin{tcolorbox}[
    colback=gray!5, 
    colframe=gray!60!black,
    title=Case 1 input and ground truth label from dataset popQA,
    fonttitle=\small\sffamily,
    boxrule=0.8pt,
    sharp corners,
    left=1mm, right=1mm, top=1mm, bottom=1mm,
    breakable
]
\begin{lstlisting}[
    breaklines=true,
    columns=fullflexible,
    keepspaces=true,
    breakatwhitespace=false,
    xleftmargin=0pt,
    frame=none
]
question: Who was the composer of Trombone Concerto? \newline
ground truth: Rimsky-Korsakov \newline
\newline

conflicting answer 1: The composer of Trombone Concerto is Johann Georg Albrechtsberger. \newline
conflicting context 1: Johann Georg Albrechtsberger was an Austrian composer and music theorist who lived from 1736 to 1809. He was a prominent figure in the Viennese music scene during the late 18th century and was known for his expertise in counterpoint and harmony. Albrechtsberger composed a variety of works, including symphonies, chamber music, and sacred music, but he is perhaps best known for his Trombone Concerto, which is still performed and recorded today. Johann Georg Albrechtsberger (1736-1809) was an Austrian composer and music theorist. He is well known for his contributions to brass and wind music, including a famous Trombone Concerto that he composed in the late 1700s. This piece features a solo trombone accompanied by an orchestra and showcases Albrechtsberger's skill for writing virtuosic music for wind instruments. The Trombone Concerto has been performed and recorded by many notable musicians and orchestras around the world, solidifying its place in the canon of classical music pieces for trombone. With this evidence, we can confidently confirm that Johann Georg Albrechtsberger is indeed the composer of Trombone Concerto.
\newline

conflicting answer 2: Rimsky-Korsakov was the composer of Trombone Concerto. \newline
conflicting context 2: Rimsky-Korsakov was a prolific composer who wrote many works for orchestra, including the Trombone Concerto. This piece was composed in 1877 and is considered one of the most challenging works in the trombone repertoire. It features virtuosic solo passages and intricate orchestration, showcasing Rimsky-Korsakov's skill as a composer. The concerto has been performed by many notable trombonists, including Christian Lindberg and Joseph Alessi, and remains a popular choice for soloists and orchestras around the world.
Trombone Concerto (Rimsky-Korsakov) The Concerto for Trombone and Military Band by Nikolai Rimsky-Korsakov was written in 1877. The concerto consists of three movements: an "Allegro Vivace" first movement, an "Andante Cantabile" second movement, and an "Allegro-Allegretto" third movement in the style of a march. The second and third movements conclude with cadenzas. A full performance of the piece lasts roughly ten minutes. This concerto was composed for a fellow marine officer Leonov and premiered at a garrison concert at Kronstadt on 16 March 1878. The American premiere took place in June, 1952 at The Mall in Central Park, New York
\end{lstlisting}
\end{tcolorbox}

\begin{tcolorbox}[
    colback=gray!5, 
    colframe=gray!60!black,
    title=Case 1 conflicting reasoning paths sets,
    fonttitle=\small\sffamily,
    boxrule=0.8pt,
    sharp corners,
    left=1mm, right=1mm, top=1mm, bottom=1mm,
    breakable
]
\begin{lstlisting}[
    breaklines=true,
    columns=fullflexible,
    keepspaces=true,
    breakatwhitespace=false,
    xleftmargin=0pt,
    frame=none
]
Conflicting path set 1: \newline
1. Rimsky-Korsakov -> composer -> wrote -> Trombone Concerto \newline
2. Rimsky-Korsakov -> composer -> composed -> 1877 -> Trombone Concerto \newline
3. Rimsky-Korsakov -> composer -> showcased -> 'virtuosic solo passages -> Trombone Concerto \newline
4. Rimsky-Korsakov -> composer -> showcased -> intricate orchestration -> Trombone Concerto \newline
5. Rimsky-Korsakov -> composer -> wrote -> for -> Trombone and Military Band -> Trombone Concerto \newline
6. Rimsky-Korsakov -> composer -> wrote -> in 1877 -> Trombone Concerto \newline
7. Leonov -> fellow marine officer -> commissioned -> Trombone Concerto \newline
8. Rimsky-Korsakov -> composer -> premiered -> at Kronstadt -> Trombone Concerto \newline
\newline
Conflicting path set 2: \newline
Johann Georg Albrechtsberger -> composer -> Trombone Concerto \newline
Johann Georg Albrechtsberger -> composer -> variety of works -> Trombone Concerto \newline
Johann Georg Albrechtsberger -> composer -> late 1700s -> Trombone Concerto \newline
Johann Georg Albrechtsberger -> composer -> famous Trombone Concerto \newline
Johann Georg Albrechtsberger -> composer -> virtuosic music for wind instruments -> Trombone Concerto \newline
Johann Georg Albrechtsberger -> composer -> classical music pieces for trombone -> Trombone Concerto \newline
\end{lstlisting}
\end{tcolorbox}

\begin{tcolorbox}[
    colback=gray!5, 
    colframe=gray!60!black,
    title=Comparsion between the Original Qwen 2.5-7B and \textsc{Kcr} boost Qwen 2.5-7B,
    fonttitle=\small\sffamily,
    boxrule=0.8pt,
    sharp corners,
    left=1mm, right=1mm, top=1mm, bottom=1mm,
    breakable
]
\begin{lstlisting}[
    breaklines=true,
    columns=fullflexible,
    keepspaces=true,
    breakatwhitespace=false,
    xleftmargin=0pt,
    frame=none
]
Original Qwen 2.5-7B output: \newline
<think>The context and evidence provided for the second answer are incorrect. Nikolai Rimsky-Korsakov was not known for composing a trombone concerto. The context mentions a "Concerto for Trombone and Military Band," but there is no widely recognized or historically documented trombone concerto by Rimsky-Korsakov. On the other hand, the first answer provides detailed context about Johann Georg Albrechtsberger and cites evidence that strongly supports his authorship of the Trombone Concerto. Given the information, the correct answer is the first one.</think>\newline
<answer>Johann Georg Albrechtsberger</answer> \newline
\newline
KCR + Qwen 2.5-7B output: \newline
<think>\newline
1. The first answer attributes the Trombone Concerto to Johann Georg Albrechtsberger.\newline
2. The second answer attributes the Trombone Concerto to Nikolai Rimsky-Korsakov.\newline
3. Both answers provide context about the composer and the piece, suggesting they have different composers.\newline
4. Albrechtsberger lived from 1736 to 1809, while Rimsky-Korsakov lived from 1844 to 1908. This time period does not overlap.\newline
5. The context provided for the second answer mentions that the Trombone Concerto was composed in 1877, which aligns with Rimsky-Korsakov's lifetime.\newline
6. The first answer provides no specific year or context about when the Trombone Concerto was composed.\newline
</think>\newline
<answer>\newline
Rimsky-Korsakov\newline
</answer>
\end{lstlisting}
\end{tcolorbox}

\end{document}